\documentclass[letterpaper, 10 pt, conference]{ieeeconf}  

\IEEEoverridecommandlockouts                              
\usepackage{booktabs}
\usepackage{multirow}
\usepackage{graphicx}
\usepackage{bm}
\usepackage{cite}
\usepackage{amsmath} 

\title{\LARGE \bf
Continuous-time Gaussian Process Trajectory Generation for Multi-robot Formation via Probabilistic Inference
}

\author{Shuang Guo$^{*}$$^{1}$, 
        Bo Liu$^{*}$$^{2}$, 
        Shen Zhang$^{3}$,
        Jifeng Guo$^{1}$
        and Changhong Wang$^{2}$
\thanks{*The first two authors contributed equally to this article}
\thanks{$^{1}$Department of Aerospace Engineering, Harbin Institute of Technology, Harbin, China.
    {\tt\small shuang\_guo.robotics@outlook.com}, 
    {\tt\small guojifeng@hit.edu.cn}}%
\thanks{$^{2}$Space Control and Inertial Technology Research Center, 
Harbin Institute of Technology, Harbin, China.
    {\tt\small hitlb2017@gmail.com}, 
    {\tt\small cwang@hit.edu.cn}}%
\thanks{$^{3}$Department of Microelectronics, Harbin Institute of Technology, Harbin, China.
    {\tt\small colson\_z@outlook.com}}%
}

\begin{document}
\maketitle
\thispagestyle{empty}
\pagestyle{empty}

\begin{abstract}

In this paper, we extend a famous motion planning approach, GPMP2, to multi-robot cases, yielding a novel centralized trajectory generation method for the multi-robot formation. A sparse Gaussian Process model is employed to represent the continuous-time trajectories of all robots as a limited number of states, which improves computational efficiency due to the sparsity. We add constraints to guarantee collision avoidance between individuals as well as formation maintenance, then all constraints and kinematics are formulated on a factor graph. By introducing a global planner, our proposed method can generate trajectories efficiently for a team of robots which have to get through a width-varying area by adaptive formation change. Finally, we provide the implementation of an incremental replanning algorithm to demonstrate the online operation potential of our proposed framework. The experiments in simulation and real world illustrate the feasibility, efficiency and scalability of our approach.

\end{abstract}
\section{INTRODUCTION}

Multi-robot teams have been popularized in a wide range of tasks, including surveillance, inspection and rescue. The multi-robot team is required to move in a proper formation in some scenarios, for instance, to survey an area collaboratively \cite{RN11}. Trajectory generation is an indispensable component in multi-robot systems \cite{RN6}. It is challenging for planning algorithms to efficiently compute the goal-oriented, collision-free trajectories while respecting kinematics and formation constraints because of a large number of robots sharing the same space\cite{RN5}.

Current multi-robot motion planning methods can be classified into two categories, namely decentralized methods and centralized methods \cite{jackson2020scalable}. In decentralized methods \cite{RN5}, local interactions between neighbors are employed to achieve group behaviors, so decentralized methods have attracted much attention due to the reduced communication requirements and scalability. However, it is hard for them to impose constraints at either the individual or system level. By comparison, centralized approaches \cite{RN14}\cite{RN13} provide global guarantees and are reasonable about constraints, but they often scale poorly with the growing number of robots. In this paper, we present a centralized method with good scalability. Its computational cost increases cubically with the size of state\cite{RN2}, rather than exponentially as described in other literatures. We demonstrate that our method can compute the whole trajectories for 10 robots within 0.39s in a complex task where a multi-robot team is required to get through a width-varying area in formation.

\begin{figure}[t]
    \centering
    \includegraphics[scale=0.41]{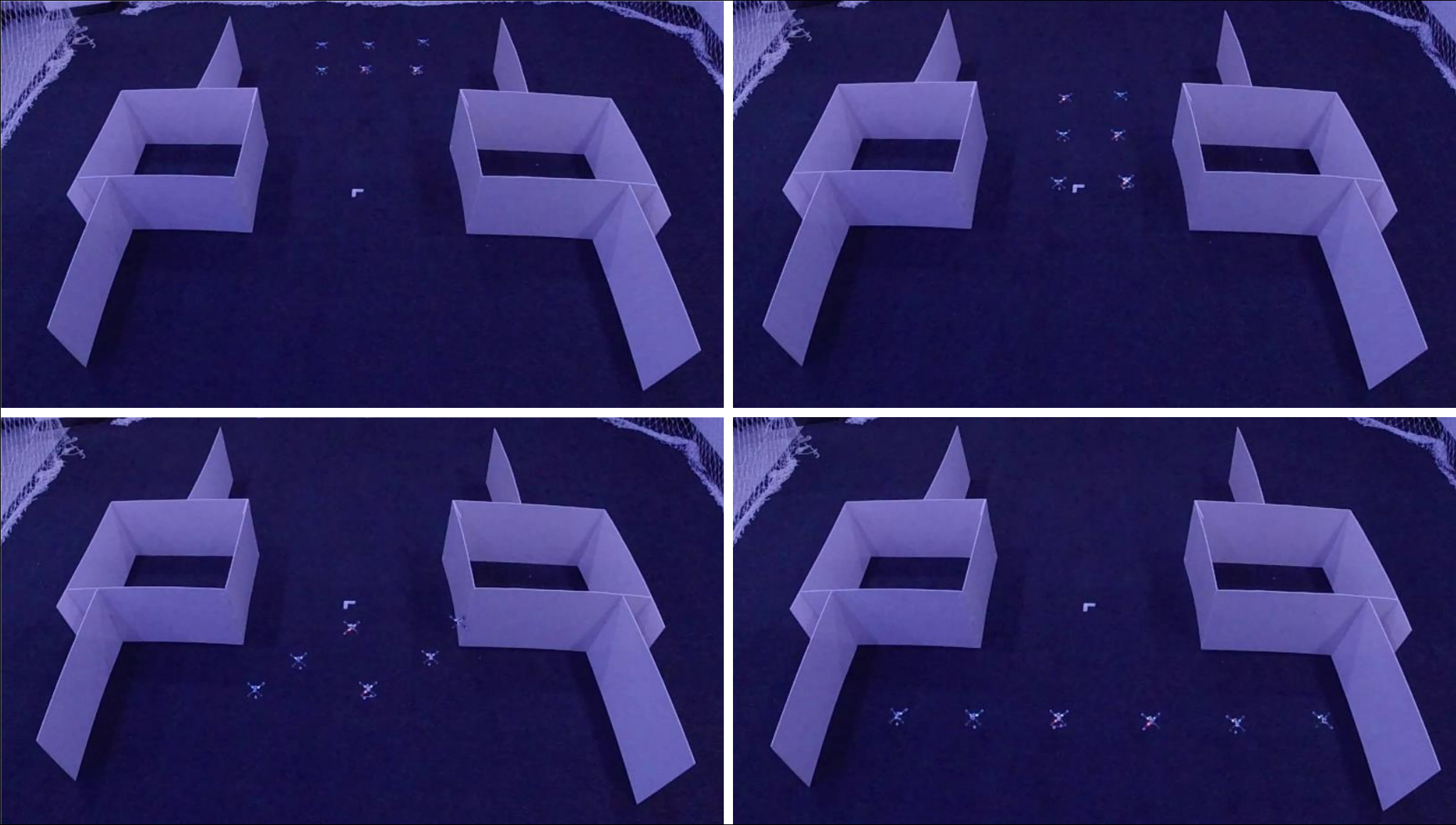}
    \caption{6 quadrotors move through a width-varying area by changing formation adaptively.}
    \label{Fig1}
\end{figure}

There are also a large body of existing works to address formation control and trajectory generation problems, including methods using reactive behaviors \cite{balch1998behavior}, potential fields \cite{balch2000social}, virtual structures \cite{zhou2018agile}, leader follower \cite{RN10} and model predictive control \cite{dunbar2002model}.  Some other researchers have formulated the multi-robot formation navigation problem as a constrained optimization. In \cite{RN11}\cite{RN14}, a sequential convex programming is used to navigate a multi-robot formation to the goal while reconfiguring the formation to avoid obstacles. However, most existing works are limited to hold a fixed formation, or transition between several predefined formations. In contrast, given a map, our proposed method can adaptively compute proper rectangular formations and then allocate execution time intervals for every formation so that the multi-robot team can get through a width-varying area without human designers.

In this work, we extend GPMP2 \cite{RN2}, a well-known motion planning algorithm using Gaussian Processes (GPs) and factor graphs, to the multi-robot formation case. We represent continuous-time trajectories of all robots in the formation as samples from a GP and then formulate the trajectory optimization problem as probabilistic inference expressed on a factor graph, which can be solved fast by exploiting the sparsity. Constraints on kinematics, obstacle avoidance, collision avoidance between individuals as well as formation maintenance are all formulated as factors deployed on the factor graph to ensure that the feasible trajectory for every robot in the team can be found by performing a non-linear least-square optimization. 

The multi-robot team needs the ability to change the formation adaptively when moving through a width-varying area, as shown in Fig. \ref{Fig1}. To this end, we introduce a global planner that includes two parts: formation planning and task assignment. Given a group of robots with known initial locations and the target point of the formation, as well as a map describing the environment, expected formations and their corresponding execution time intervals can be computed efficiently. Then a simple but effective task assignment method specific to rectangular formations is employed to allocate each robot to its unique position in the formation. In this way, conflicts during the formation transition can be avoided, which guarantees trajectory optimization to converge to the optimal solution. All these results are then used to define formation constraints that are respected in the trajectory optimization. 

In most cases, centralized approaches run offline \cite{jackson2020scalable}, which means a known global map is required in advance. However, in many tasks, only a limited sensing range around robots is available, or the destination of the formation is varying. To this end, we implement an incremental replanning algorithm for multi-robot formations following iGPMP2 \cite{RN2} to illustrate the online operation potential of our framework. 
\section{GLOBAL PLANNING}
\subsection{Formation Planning}
Inspired by \cite{Liu2018}\cite{RN25}, we use Rectangular Safe Flight Corridor (RSFC) to generate expected formations. Then we allocate the corresponding execution time interval for each formation according to lengths of all path segments. To this end, we present three procedures: $(i)$ RSFC construction, $(ii)$ formation generation, and $(iii)$ path updating and time allocation. Note that we focus on rectangular formations because they can cover most of the practical tasks.
\subsubsection{RSFC Construction}
\begin{figure}[t]
    \centering
    \includegraphics[scale=0.68]{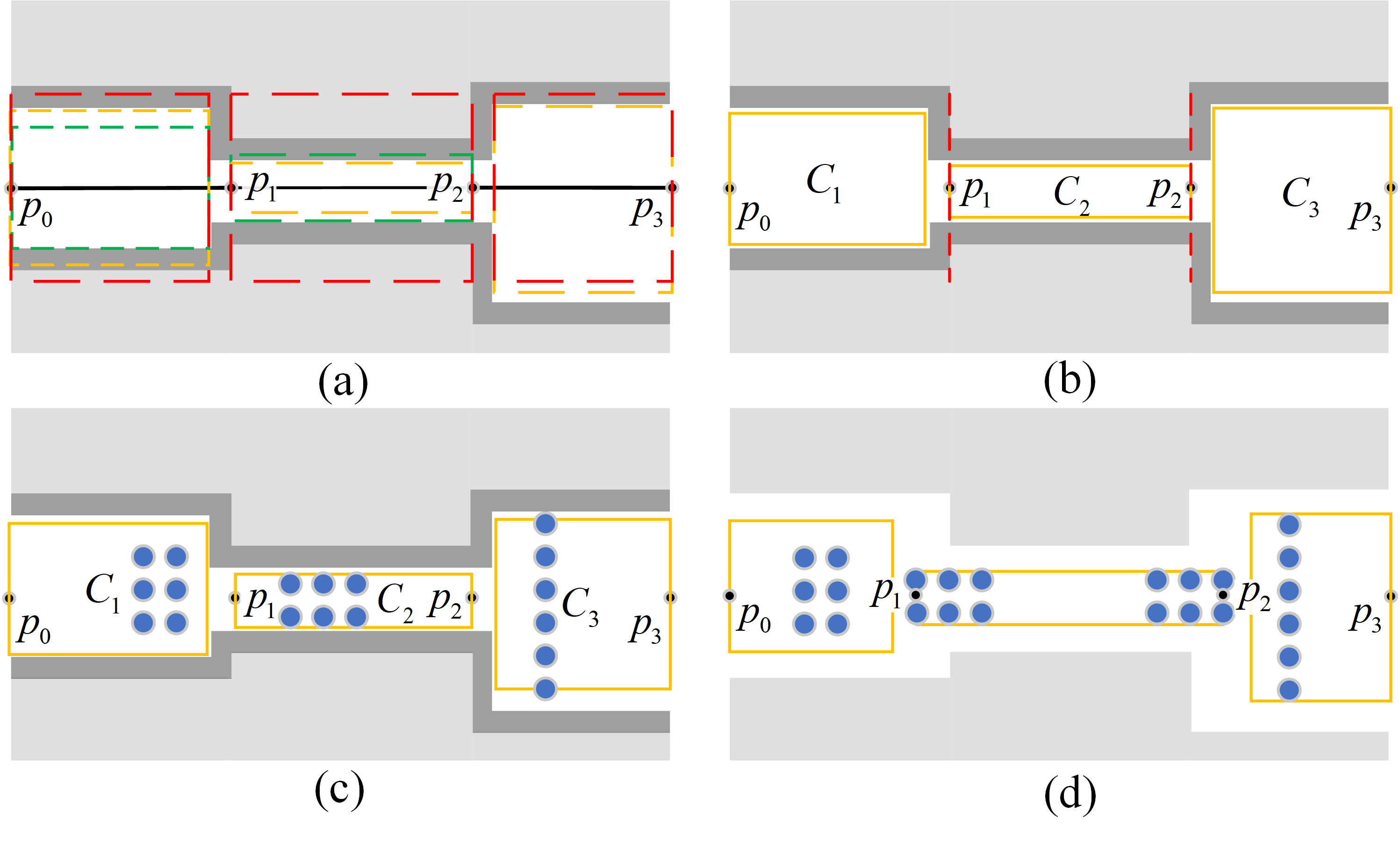}
    \caption{Formation planning in a known map: (a) initialize the RSFC (red dashed line)  and find two collision-free boundaries (yellow and green dashed line); (b) build a rectangular corridor ${C_i}$; (c) construct a formation according to ${C_i}$.  (d) update the path and the RSFC to ensure that all robots are free of collision and add a time gap for the formation transition.}
    \label{Fig2}
\end{figure}

 As shown in Fig. \ref{Fig2}, obstacles are represented by gray shaded areas. We adopt the convention used in \cite{Liu2018}. A piece-wise linear path is donated as ${P}=\langle\bf{p}$$_{0} \rightarrow \bf{p}$$_{1} \rightarrow \ldots \rightarrow \bf{p}$$_{M}\rangle$,  where $\bf{p}$$_i$ indicates a point in the free space and $\bf{p}$$_i \to \bf{p}$$_{i + 1}$ is a directed line segment, donated as ${L}_i = \langle\bf{p}$$_{i} \rightarrow \bf{p}$$_{i+1} \rangle$. In our work, $\bf{p}$$_i$ is initialized by the positions of width changes. The RSFC generated from ${L}_i$ is denoted as ${C_i}$. In this step we find a collision-free RSFC which includes the segment ${L}_i$. Robots are modeled as circles with radius of $\varepsilon $ and we expand the obstacle region with the thickness of $r \text{ }( r > \varepsilon ) $ to ensure robots obstacle-free (darkgray shaped areas in Fig. \ref{Fig2}). Then ${C_i}$ is computed in two steps: $(i)$ build an initial RSFC to derive the maximal width of the formation according to $\bf{p}$$_i$, $\bf{p}$$_{i + 1}$ and the number of robots in the team; $(ii)$ translate the initial RSFC in the direction perpendicular to ${L}_i$ or shrink it iteratively until it just fits into the obstacle-free region, so that we find ${C_i}$.
 
\subsubsection{Formation generation}
With ${C_i}$ and the expected distance between robots ${d_0}$, we can calculate the number of robots that can be accommodated in each column of the formation, which is always set to the maximal feasible value for the robot team to get through.
\subsubsection{Path updating and time allocation}
To guarantee obstacle avoidance during the formation transition, we update $\bf{p}$$_i\text{ } (i = 1,...,M - 1)$  according to the length of the desired formation (to ensure that the formation transition is carried out in the wider area, see Fig. \ref{Fig2} (d)). 
Then we allocate the corresponding execution time interval for each formation according to the path length percentage to updated ${L}_i$ in ${P}$.  
In addition, we allocate a small time gap $\tau $ for the formation transition to achieve trajectory smoothness. 
\subsection{Task Assignment}

We introduce a task assignment algorithm for rectangular formations to allocate the unique position in the formation for each robot. By doing this, conflicts between individuals during the formation transition can be significantly reduced, so it is more likely and faster for the optimizer to find the feasible trajectories. On the other hand, given the goal of the formation, the goal of each robot can be calculated automatically using the result of the task assignment. 

\begin{figure}[t]
    \centering
    \includegraphics[scale=0.45]{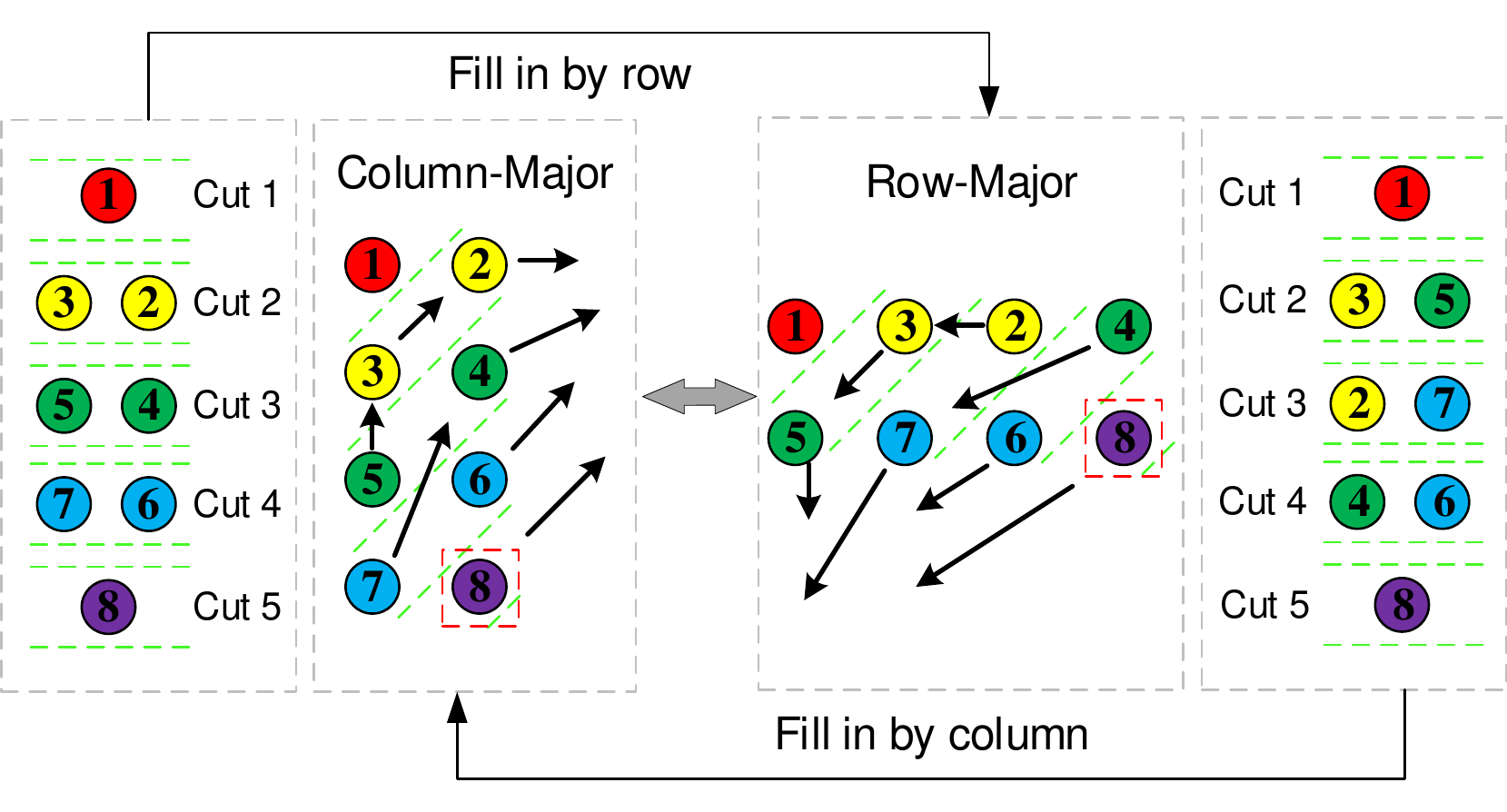}
    \caption{An example of 8 robots transitioning between 4 $\times$ 2 (column-major) and 2 $\times$ 4 (row-major) formation. All robots' relative position changes are roughly in the same direction. If the robot number changes to 7, there will be a vacancy (marked by red boxes). In this case, we first fill the vacancy with a virtual robot (robot 8) to make the rectangle full, then employ the proposed strategy to conduct position assignment.}
    \label{Fig3}
\end{figure}

During formation transitions, our strategy is to encourage the relative position changes of all robots to be roughly in the same direction. In this paper, the “row-major” formation indicates that the rectangular formation has more elements in each row than column, and “column-major” denotes the opposite. A specific example of 8 robots is shown in Fig. \ref{Fig3}. It can be seen that all robots' relative position changes are roughly in the same direction in both two cases (lower left for the case of “row-major” to “column-major” and upper right for the opposite). By doing this, all robots can reach their expected positions in the new formation in a conflict-free style. In the implementation, a matrix is used to store the position assignment scheme of each formation. Taking the transition from “column-major” to “row-major” as an example, we iteratively cut elements in the previous formation with diagonal lines (marked in green dashed lines) from top to bottom, and store them in a queue in ascending order by column indexes, then fill them into the matrix representing the new formation by row. Note that if the number of elements in the queue is more than that of vacancies in the current row of the new formation (e.g. Cut 3 in Fig. \ref{Fig3}), we first fill the remaining vacancies with the elements at the end of the queue, then use the elements left at the front of the queue to fill in a new row. The opposite is true in the inverse process. It can be proven that our method can adapt to any rectangular formation. If the robot number cannot be exactly divided by the expected formation width, there may be some vacancies in the formation. In this case, we first create virtual robots in the vacancies to make the rectangle full, in order to ensure that the proposed assignment strategy still works. When we use elements in the queue to fill in the new formation, virtual robots will be skipped.

\section{TRAJECTORY OPTIMIZATION}

Our work builds upon a well-developed motion planning algorithm GPMP2 \cite{RN2}. For the sake of completeness, we briefly review GPMP2 first. Then we introduce the new constraints we add in order to extend it to the multi-robot formation case. Finally, we also provide an incremental replanning method for the multi-robot formation by following iGPMP2  \cite{RN2} to show the online operation potential of our proposed centralized method.
\subsection{Review of GPMP2}
\subsubsection{Planning as inference on factor graphs}
GPMP2 treats the motion planning problem as probabilistic inference. The goal is to find the maximum a posterior (MAP) trajectory given a prior distribution on the space of trajectories encouraging smoothness and a likelihood function that encourages the trajectory to be collision-free \cite{RN7}, as shown in
\begin{equation}
    {{\boldsymbol{\theta }}^ * } = \mathop {\arg \max }\limits_{\boldsymbol{\theta }} P\left( {{\boldsymbol{\theta }}|{\bf{e}}} \right) 
    \label{MAP}
\end{equation}
where ${\bf{e}}$ is a set of random binary events of interest, for example, obstacle avoidance. The posterior distribution of ${\boldsymbol{\theta }}$ given ${\bf{e}}$ can be derived from the prior and likelihood by Bayes rule
\begin{equation}
    P\left( {{\boldsymbol{\theta }}|{\bf{e}}} \right) \propto P\left( {\boldsymbol{\theta }} \right)L\left( {{\bf{e}}|{\boldsymbol{\theta }}} \right)
    \label{bayes}
\end{equation}
which can be represented as the product of a series of factors
\begin{equation}
    P\left( {{\boldsymbol{\theta }}|{\bf{e}}} \right) \propto \prod\limits_{m = 1}^M {{f_m}\left( {{{\boldsymbol{\Theta }}_m}} \right)} 
    \label{factors}
\end{equation}
where ${f_m}$ are factors on state subsets ${{\boldsymbol{\Theta }}_m}$. It is shown in \cite{RN7} that this MAP problem can be expressed on a factor graph and solved in high efficiency by exploiting sparsity.
\subsubsection{The GP prior}
 A vector-valued Gaussian Process (GP) is employed to represent a continuous-time trajectory:  
 $\boldsymbol{\theta}(t) \sim \mathcal{G} \mathcal{P}\left(\boldsymbol{\mu} (t), {\bf{K}}\left(t, t^{\prime}\right)\right)$, 
 where ${\boldsymbol{\mu }}\left( t \right)$ is the mean and ${\bf{K}}\left( {t,t'} \right)$ is the covariance, which is generated by a linear time-varying stochastic differential equation (LTV-SDE) defined as
 \begin{equation}
     {\boldsymbol{\dot \theta }}\left( t \right) = {\bf{A}}\left( t \right){\boldsymbol{\theta }}\left( t \right) + {\bf{u }}\left( t \right) + {\bf{F}}\left( t \right){\bf{w}}\left( t \right)
     \label{LTV-SDE}
 \end{equation}
 where ${\bf{A}}\left( t \right)$ and ${\bf{F}}\left( t \right)$ are system matrices, $ \bf{u} \left( t \right)$ is the control input and the white noise is ${\bf{w}}\left( t \right) \sim {\cal G}{\cal P}\left( {{\bf{0}},{{\bf{Q}}_c}\delta \left( {t-t'} \right)} \right)$ with ${{\bf{Q}}_c}$ being the power-spectral density matrix and $\delta \left( {t-t'} \right)$ being the Dirac delta function. The first order moment (mean) and second order moment (covariance) can be derived from the solution to (\ref{LTV-SDE}), given by
\begin{equation}
     \widetilde{\boldsymbol{\mu}}(t)=\mathbf{\Phi}\left(t, t_{0}\right) \boldsymbol{\mu}_{0}+\int_{t_{0}}^{t} \mathbf{\Phi}(t, s) \mathbf{u}(s) \mathrm{d} s
     \label{eq6}
\end{equation}
\begin{equation}
\begin{array}{rl}
\widetilde{\bf{K}}\left( {t,t'} \right)=&{\bf{\Phi }}\left( {t,{t_0}} \right){{\bf{K}}_0}{\bf{\Phi }}{\left( {t',{t_0}} \right)^ \top } +\\
&\int_{{t_0}}^{\min \left( {t,t'} \right)} {\bf{\Phi }} (t,s){\bf{F}}(s){{\bf{Q}}_c}{\bf{F}}{(s)^ \top }{\bf{\Phi }}{\left( {t',s} \right)^ \top }{\rm{d}}s
\end{array}
\label{eq7}
\end{equation}
where ${\bf{\Phi }}$ is the state transition matrix and ${{\boldsymbol{\mu }}_0},{\rm{ }}{{\bf{K}}_0}$ are respectively mean and covariance at ${t_0}$. The Markov property of (\ref{LTV-SDE}) results in the sparsity of the inverse kernel matrix ${{\bf{K}}^{ - 1}}$ which allows for fast inference. The proof of the sparsity can be found in \cite{RN17}.
Then the GP prior can be written as 
\begin{equation}
    P({\boldsymbol{\theta }}) \propto \exp \left\{ { - \frac{1}{2}\left\| {{\boldsymbol{\theta }} - {\boldsymbol{\mu }}} \right\|_{\bf{K}}^2} \right\}
    \label{GPprior}
\end{equation}
\subsubsection{The likelihood function}
GPMP2 formulates constraints as events that the trajectory has to obey. For example, the likelihood function of obstacle avoidance indicates the probability of being free from collisions with obstacles. All likelihood functions are defined as a distribution in the exponential family, given by
\begin{equation}
    L({\boldsymbol{\theta }};{\bf{e}}) \propto \exp \left\{ { - \frac{1}{2}\left\| {{\bf{h}}({\boldsymbol{\theta }})} \right\|_{\bf{\Sigma }}^2} \right\}
    \label{eq9}
\end{equation}
where $\bf{h}\left( {\boldsymbol{\theta }} \right)$ is a vector-valued cost function and $\bf{e}$ is the corresponding events. For the proof of sparsity of the likelihood in GPMP2, please see \cite{RN2}.

\subsubsection{MAP inference}
Using (\ref{bayes}) (\ref{GPprior}) (\ref{eq9}), the MAP problem can be formulated as 
\begin{equation}
    {{\boldsymbol{\theta }}^*} = \mathop {{\mathop{\rm argmin}\nolimits} }\limits_{\boldsymbol{\theta }} \left\{ {\frac{1}{2}\left\| {{\boldsymbol{\theta }} - {\boldsymbol{\mu }}} \right\|_{\bf{K}}^2 + \frac{1}{2}\left\| {{\bf{h}}({\boldsymbol{\theta }})} \right\|_{\bf{\Sigma }}^2} \right\}
\end{equation}
which is a well-studied non-linear least square problem. Therefore, the optimal trajectory can be found by solving it using iterative algorithms such as Gauss-Newton method and Levenberg-Marquardt (L-M) method. 
\subsection{Likelihood specific to the multi-robot formation case}
\subsubsection{Formation Constraints}
On the basis of GPMP2, we additionally include a formation constraint on the factor graph to enforce the robot team to maintain the expected formation.

\begin{figure}[t]
    \centering
    \includegraphics[scale=0.8]{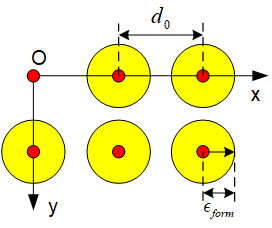}
    \caption{The definition of the formation coordinate: ${d_0}$ is the expected distance between robots and $\epsilon_{form}$ is the radius of the allowable range for each robot.}
    \label{Fig4}
\end{figure}

Our definition of the formation is depicted in Fig. \ref{Fig4}. The robot in the upper left corner is the origin point of the formation. The red points indicate the expected position of each robot, which is computed by the expected distance   and the relative position to the origin point. The tolerant range of each robot is marked as yellow circles with the radius $\epsilon_{form}$ in Fig. \ref{Fig4}. We achieve formation control by limiting the relative position of each robot to the origin. In contrast to giving an expected global position to every robot and controlling each alone to its goal, our strategy enables the robot team to act more flexibly, and their behaviors would be closer to a real intelligent swarm, instead of a set of individuals executing their own orders respectively.
\subsubsection{Collision avoidance between individuals}
Another constraint required in the multi-robot case is collision avoidance between each other during the movement, especially during the process of the formation transition.

We prevent collisions between each other by checking the distances between every two robots in the multi-robot team. If any two robots get too close to be safe, it will cause a rapid increase in the value of the corresponding cost function. In this way, robots will show mutual repulsion and keep a safe distance with each other.
\subsection{Factor graph formulation}
\begin{figure}[t]
    \centering
    \includegraphics[scale=0.45]{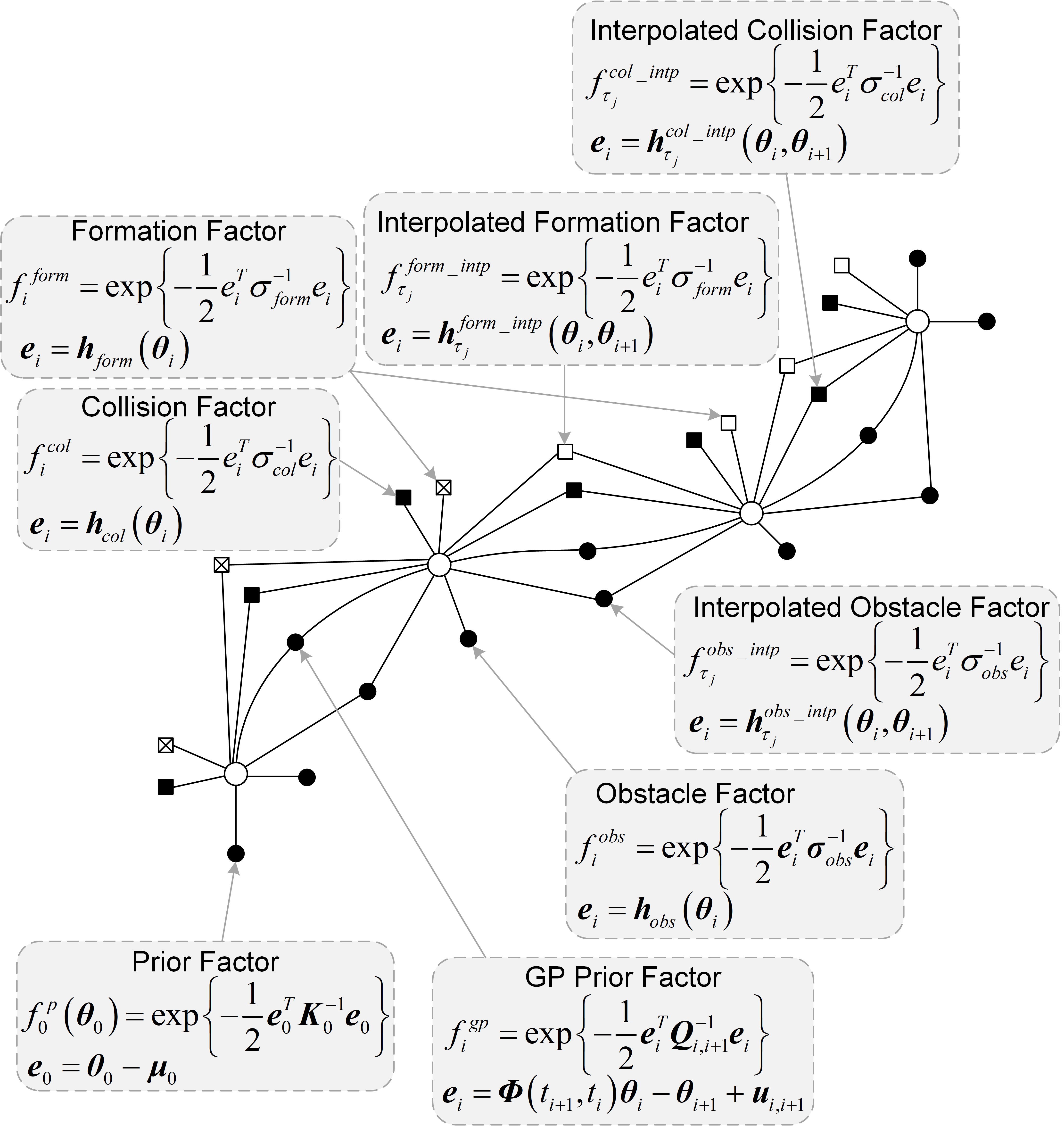}
    \caption{A factor graph of an example trajectory optimization problem for a multi-robot formation. The support states are marked as white circles and 4 types of factors (namely prior factors on the start and goal states, GP prior factors, obstacle factors and the corresponding interpolated versions between consecutive support states) which have already been implemented in GPMP2 are marked as black dots. The factors marked as squares are specific to the multi-robot formation, namely collision factors (black squares), formation factors (white squares) and their corresponding interpolated versions. Note that some white squares have cross marks, which means we can impose different formation constraints on states of different time, so that the adaptive formation change can be achieved.}
    \label{Fig5}
\end{figure}

Having defined all factors, we now describe our graphical model for representing the problem of trajectory optimization for a multi-robot formation. An example is illustrated in Fig. \ref{Fig5}. Compared with GPMP2, we add two types of unary factors: formation factors and collision factors\footnote{Note that collision factors here refer specifically to collisions between robots, those of static obstacle avoidance are indicated by obstacle factors.}, and their corresponding interpolated versions to the factor graph. Therefore, the useful sparsity exploited by GPMP2 is still available in our case. A simple explanation for interpolated factors is that we can compute the state of any time of interest between two support states and impose constraints on it (For more details about the GP interpolation, see \cite{RN2}\cite{RN20}\cite{RN21}). Furthermore, this constraint can be equivalently allocated to two support states due to the property of GPs (Fig. \ref{Fig5}). It is worth noting that we apply two different formation factors on the trajectories in Fig. \ref{Fig5}, meaning that we are allowed to use different formation configurations at any time of interest. In our work, we define the formation factors according to the result of the global planning, which includes a sequence of expected formations and their corresponding time intervals of execution.
\subsection{Incremental inference for replanning}
We also provide the implementation of an incremental replanning algorithm on the basis of iGPMP2, which is necessary when the target point of the formation have been moved or new obstacles are found by robots due to the limited sensing scope.

Given a set of optimized trajectories and the changed condition, the replanning task is to efficiently recompute new feasible trajectories for the robot team to achieve its goal on the premise of safety. Following iGPMP2, we adopt an incremental style to update the current solution by using the Bayes Tree \cite{RN19}\cite{RN18} data structure, instead of resolving a new entire MAP inference from scratch. By doing this, we can update trajectories fast to achieve the online operation, because the main body of the original problem is unchanged. For a full treatment about how the incremental method works, see \cite{RN2}. 

In the implementation, we use our proposed framework to solve the original trajectory generation problem. Then we update the factor graph according to the changed conditions and an incremental solver called iSAM2 \cite{RN18} is adopted to update the trajectories for all robots in the formation. 
\section{IMPLEMENTATION DETAILS}

We implement our global planner in C++ from scratch. The C++ implementation of our trajectory optimization part in the proposed framework builds upon open-source libraries: GPMP2 \cite{RN2} and GTSAM \cite{RN22}. Following GPMP2, we employ Levenberg-Marquardt algorithm to solve the non-linear least square problem and the default parameters are adopted. Similarly, our replanning implementation adopts iSAM2 incremental solver with default settings. It is worth noting that our trajectory optimizers are initialized by a constant-velocity straight line trajectory from the start to the goal for every robot, without any prior about the environment. Since it is straightforward for our trajectory optimization part to be extended to 3D cases, here we discuss 2D cases for brevity.
\subsection{GP prior}

We augment the state variable to include all robots' positions and velocities
\begin{equation}
    {\boldsymbol{\theta }}\left( t \right) = {\left[ {\begin{array}{*{20}{c}}
{{{\bf{x}}_1}\left( t \right)}& \cdots &{{{\bf{x}}_N}\left( t \right)}&{{{{\bf{\dot x}}}_1}\left( t \right)}& \cdots &{{{{\bf{\dot x}}}_N}\left( t \right)}
\end{array}} \right]^ \top }
\label{eq11}
\end{equation}
where ${\mathbf{x}} _i \left( t \right)$ is the position of the $i$-$th$ robot in the group and $N$ is the number of robots.

Similar to GPMP2, we adopt a “constant velocity” prior model, then the LTV-SDE in (\ref{LTV-SDE}) is given by
\begin{equation}
    {\bf{A}}\left( t \right) = \left[ {\begin{array}{*{20}{c}}
{\bf{0}}&{\bf{I}}\\
{\bf{0}}&{\bf{0}}
\end{array}} \right],{\bf{u}}\left( t \right) = {\bf{0}},{\bf{F}}\left( t \right) = \left[ {\begin{array}{*{20}{c}}
{\bf{0}}\\
{\bf{I}}
\end{array}} \right]
\label{eq12}
\end{equation}
This prior model implies that our smoothness of trajectories is defined by minimizing the accelerations of all robots.
\subsection{Obstacle avoidance likelihood}

Similar to GPMP2, we compute a signed distance field (SDF) \cite{RN23} with the map which indicates obstacles in the environment. Then we use the SDF to check the distance to the closest obstacle for every robot and impose a hinge loss
\begin{equation}
    {c_{obs}}(z,{s_i}) = \left\{ {\begin{array}{*{20}{c}}
{ - {d_o}\left( {z,{s_i}} \right) + {\epsilon_{obs}}}&{{\rm{ if }\;}{d_o} < {\epsilon_{obs}}}\\
0&{{\rm{ if }\;}{d_o} \ge {\epsilon_{obs}}}
\end{array}} \right.
\end{equation}
where $d_o \left( {z,{s_j}} \right)$ is the distance to the closest obstacle for each robot ${s_i}{\rm{ }}\left( {i = 1,...,N} \right)$. Then we have the cost function
\begin{equation}
    {{\bf{h}}_{obs}}\left( {{{\boldsymbol{\theta }}_k}} \right) = {\left. {\left[ {{c_{obs}}\left( {{{\boldsymbol{\theta }}_k},{s_i}} \right)} \right]} \right|_{1 \le i \le N}}
    \label{eq15}
\end{equation}
\subsection{Collision avoidance likelihood}

We check the distances between every two robots at each iteration and impose a hinge loss on them, given by
\begin{equation}
    {{{c}}_{col}}(z,{s_i},{s_j}) = \left\{ {\begin{array}{*{20}{c}}
{ - {d_c}\left( {z,{s_i},{s_j}} \right) + {\epsilon_{col}}}&{{\rm{ if }\;}{d_c} < {\epsilon_{col}}}\\
0&{{\rm{ if }\;}{d_c} \ge {\epsilon_{col}}}
\end{array}} \right.
\label{eq16}
\end{equation}
where ${d_c}\left( {z,{s_i},{s_j}} \right)$ indicates the distance between robot $s_i$ and $s_j$. The cost function can be written as 
\begin{equation}
    {{\bf{h}}_{col}}\left( {{{\boldsymbol{\theta }}_k}} \right) = {\left. {\left[ {{c_{col}}\left( {{{\boldsymbol{\theta }}_k},{s_i},{s_j}} \right)} \right]} \right|_{1 \le i < j \le N}}
    \label{eq17}
\end{equation}
\subsection{Formation constraints}

We again employ a hinge loss to encourage robots to hold the expected formation when required.
\begin{equation}
    {c_{form}}(z,{s_i}) = \left\{ {\begin{array}{*{20}{c}}
0&{{\rm{ if }\;}{d_f} \le {\epsilon_{form}}}\\
{{d_f}\left( {z,{s_i}} \right) - {\epsilon_{form}}}&{{\rm{ if }\;}{d_f} > {\epsilon_{form}}}
\end{array}} \right.
\label{eq18}
\end{equation}
where ${d_f}\left( {z,{s_i}} \right)$ is the distance between robot ${s_i}$ and its expected position. Then we have the cost function
\begin{equation}
    {{\bf{h}}_{form}}\left( {{{\boldsymbol{\theta }}_k}} \right) = {\left. {\left[ {{c_{form}}\left( {{{\boldsymbol{\theta }}_k},{s_i}} \right)} \right]} \right|_{1 < i \le N}}
    \label{eq19}
\end{equation}
So far, we have defined all the cost functions depicted in Fig. \ref{Fig5}. The weight of each term is defined by ${\sigma _{obs}},{\sigma _{col}},{\sigma _{form}}$. Generally, a smaller $\sigma$ correlates to a higher weight.

\begin{table}[t]
\caption{Parameter settings: A higher $\sigma $ indicates a lower weight}
\label{table1}
\begin{center}   
\begin{tabular}{cccc}  
  \toprule   
    Robot number & $\sigma_{obs}$ & $\sigma_{col}$ & $\sigma_{form}$\\  
  \midrule   
    4 & 0.1 & 0.1 & 0.3   \\  
    6 &0.4 & 0.4 & 0.02    \\
    10 &0.4 & 0.4 & 0.005    \\
  \bottomrule  
\end{tabular}
\end{center}
\end{table}

\section{EXPERIMENT}
We test the proposed framework with a team of quadrotors on three common scenarios: formation maintenance, replanning for a changed destination and adaptive formation change for moving through a width-varying area. Experiments in the real world are conducted with a group of Crazyflie nano-quadrotors flying under the supervision of a NOKOV motion capture system.

The global planning is applied to the cases involved formation change. The expected distance between quadrotors in the formation is $d_0=0.5m$. The thickness of the expansion in the obstacle area is $r = 0.3m$ and the time gap is $\tau  = 2s$.

As for the trajectory optimization, all GP trajectories are represented by 11 support states, and all tasks are required to be finished in the same time of $10s$. Quadrotors are modeled as point robots with radius of $5cm$. Some other parameter settings are as follows: ${\mathbf{Q}}_c = \bf{I}$, $\epsilon_{obs}=\epsilon_{col}=0.2m$, $\epsilon_{form}=0.01m$. The parameters indicating the weight of each term of likelihood are shown in TABLE \ref{table1}.
It can be seen that the priority of formation constraints is much higher in the cases involved formation change. It implies that the formation configurations provided by the global planning are strong priors encouraging the trajectory optimization to achieve the optimal solution.

Since our work focuses on the trajectory generation problem, the output of our method is a set of trajectories represented by a sequence of discrete-time states. Therefore, we employ a post-process to convert trajectories to a set of polynomial curves with degree of 7, which can be deployed on Crazyflies. We use the CVXOPT package in Python to solve the quadratic programming. During experiments, converted trajectories are uploaded to Crazyflies in real time and executed by using Crazyswarm infrastructure support \cite{RN26}. Crazyflies receive messages from the motion capture system through CrazyRadio PA for localization.

\subsection{Formation maintenance of multiple quadrotors}
\begin{figure}[t]
    \centering
    \includegraphics[scale=0.35]{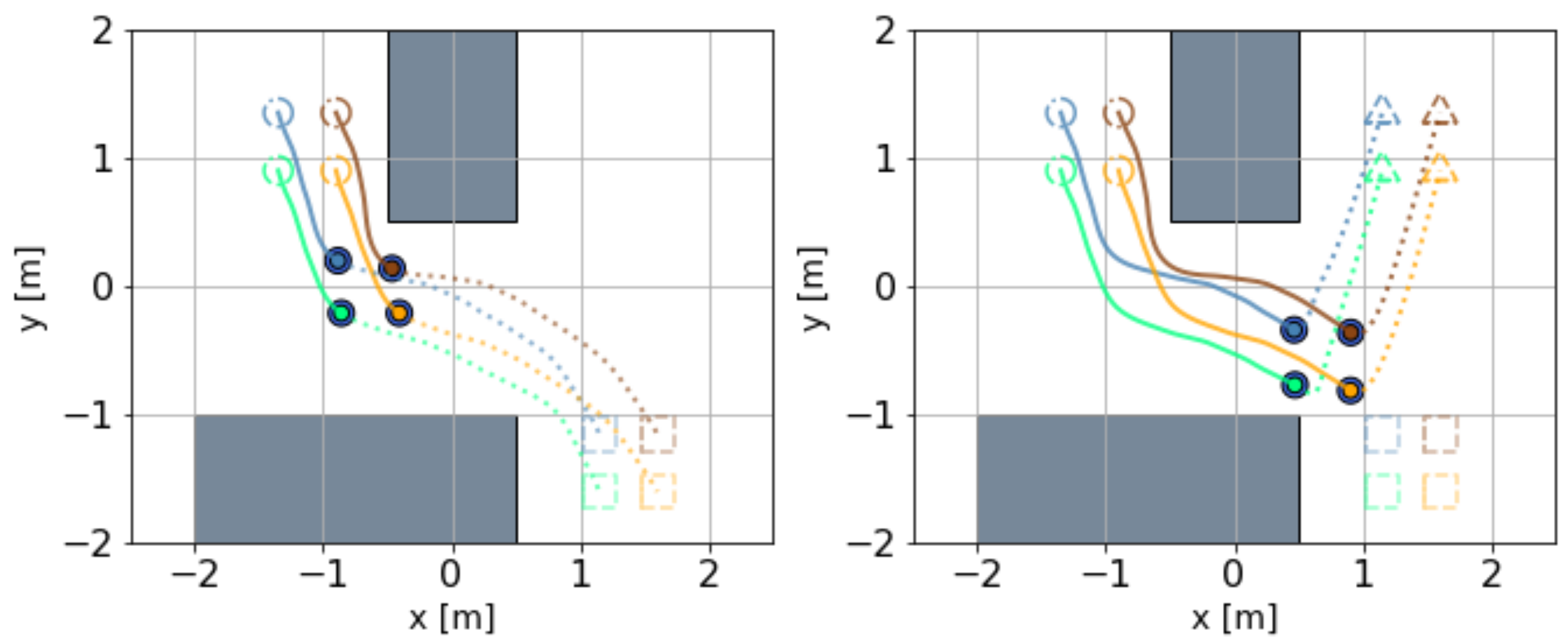}
    \caption{4 quadrotors fly to the goal while holding the square formation, as well as the corresponding goal-changed replanning case. The original goals are marked as squares and the new goals are marked as triangles. Snapshots at time 2.4s, 7.0s.}
    \label{Fig7}
\end{figure}

In the first scenario, 4 quadrotors are expected to fly to the goal in a fixed square formation, while avoiding obstacles, see Fig. \ref{Fig7}. It is worth noting that the quadrotor team makes trade-offs when necessary. In our case, the weight of formation constraints is smaller compared with the other two terms, so the formation has slight distortion during the turn but a smoother turn is achieved.

\begin{figure}[t]
    \centering
    \includegraphics[scale=0.35]{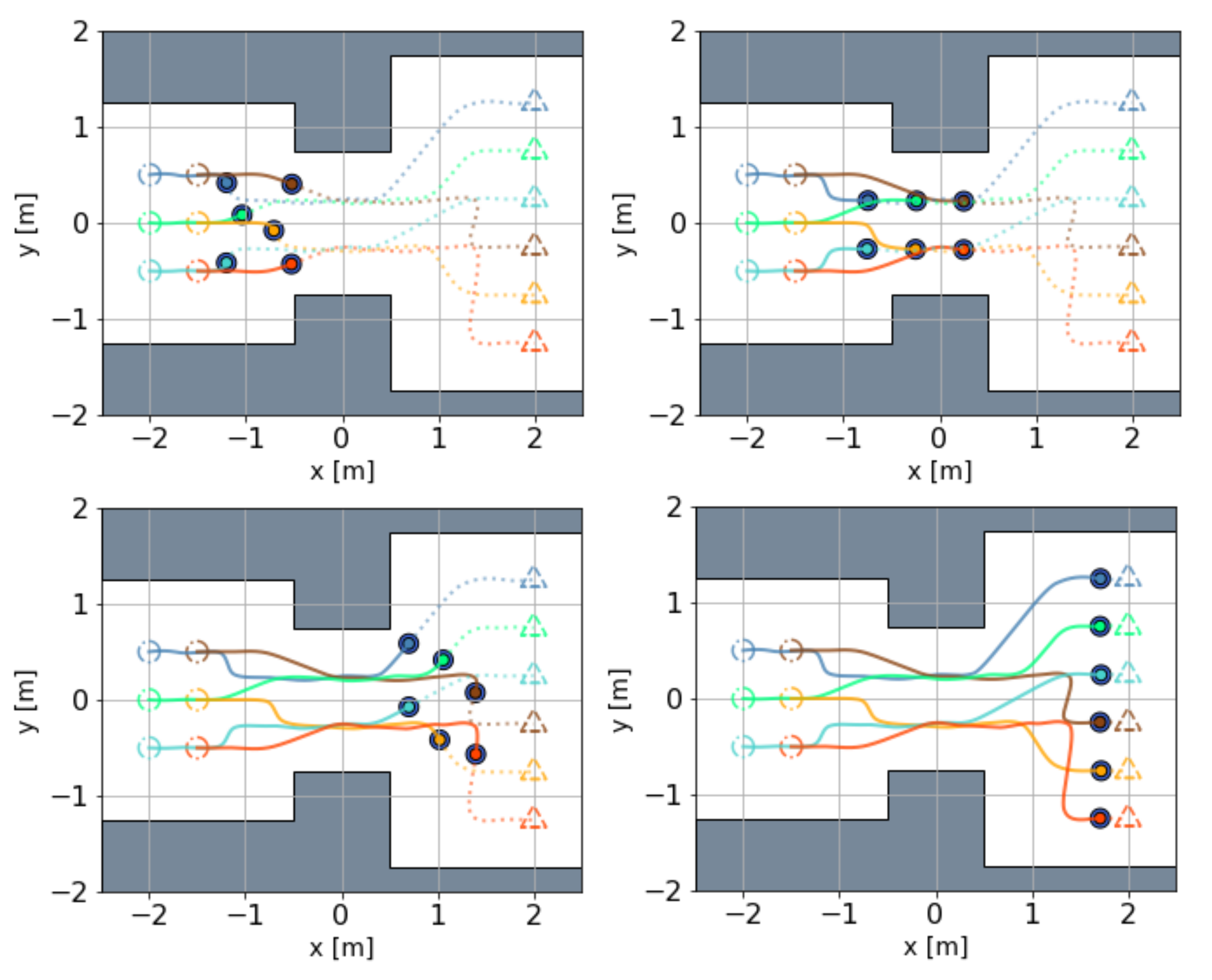}
    \caption{6 quadrotors fly through a width-varing area while adaptively changing the formation. Snapshots at time 2.4s, 4.0s, 7.4s and 9.0s.}
    \label{Fig8}
\end{figure}

\subsection{Replanning for a changed destination}
We then extend the first scenario by modifying the target point suddenly while quadrotors are flying to their original goals. We move the target point in the opposite direction to the original target at $t=7s$, as illustrated in Fig. \ref{Fig7}. Our incremental replanning algorithm updates the trajectory to the new goal within 4ms, which meets the real-time requirement.

\subsection{Adaptive formation change for a width-varying area}
\begin{figure}[t]
    \centering
    \includegraphics[scale=0.35]{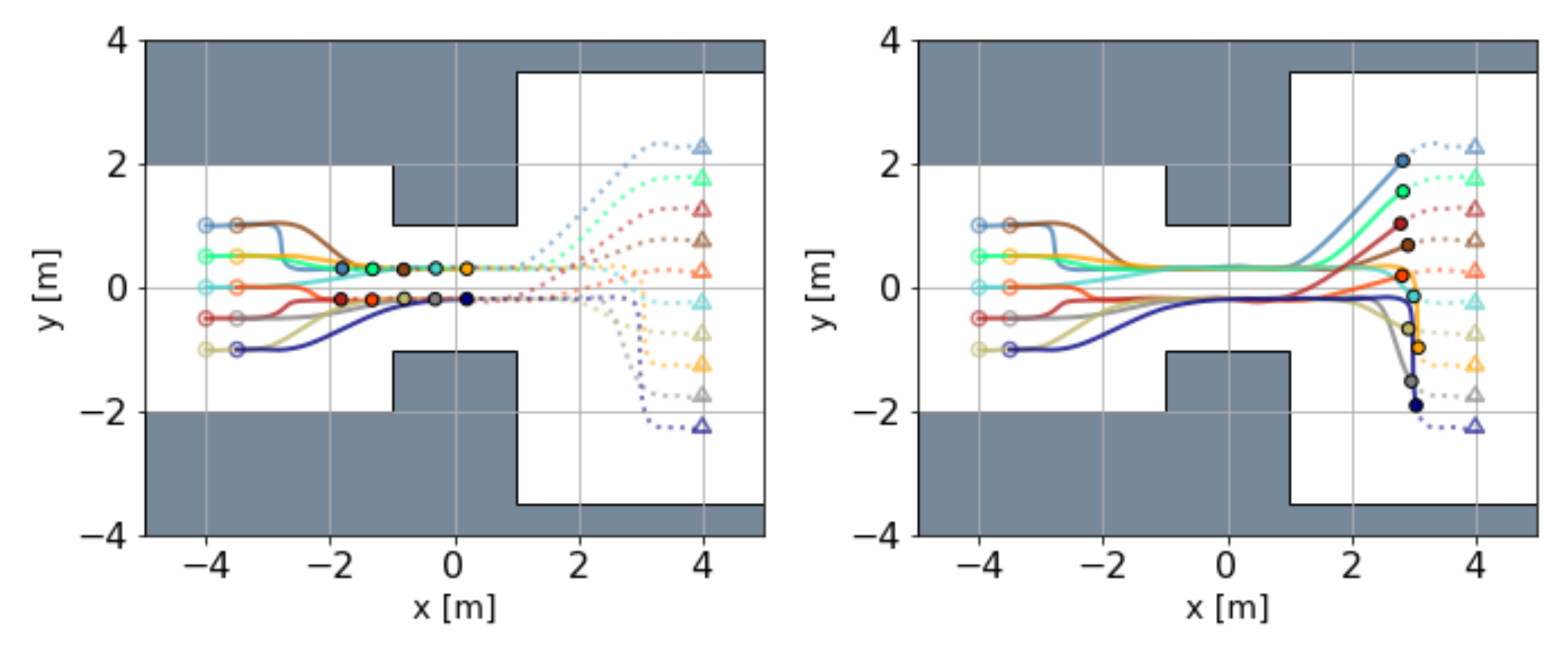}
    \caption{10 quadrotors fly through a width-varing area while adaptively changing the formation. Snapshots at time 4.0s and 8.0s.}
    \label{10robot}
\end{figure}

In the last scenario, as depicted in Fig. \ref{Fig8}, 6 quadrotors are set to move through a corridor with three different widths: 2.5m, 1.5m, 3.5m, the corresponding formations and execution time intervals calculated by the global planner are $(i)$ $1s{\rm{ }} - {\rm{ }}2s:{\rm{ }}3 \times 2$; $(ii)$ $4s{\rm{ }} - {\rm{ }}7s:{\rm{ }}2 \times 3$; $(iii)$ $9s{\rm{ }} - {\rm{ 10}}s:{\rm{ 6}} \times 1$. In addition, we implement a demo of 10 quadrotors on a four-fold map, see Fig. \ref{10robot}. The three widths are respectively: 4m, 2m, 7m, and the corresponding formation configurations are: $(i)$ $1s-2s:$ $5 \times 2$; $(ii)$ $4s-7s:$ $2 \times 5$; $(iii)$ $9s-10s:$ $10 \times 1$.

\begin{table}[t]
\caption{Results of computational time evaluation [ms]}
\label{table2}
\begin{center}   
\begin{tabular}{ccccc}  
  \toprule   
    Robot & Formation & Task & Trajectory & \multirow{2}{*}{Total} \\ 
    number & planning & assignment & optimization&  \\
  \midrule
    4 & - & - & 32.2415 & 32.2415 \\
    6 &0.0383 & 0.0149 & 75.0883 & 75.1415    \\
    10 &0.0390 & 0.0165 & 386.9926& 387.0481    \\
  \bottomrule  
\end{tabular}
\end{center}
\end{table}

\subsection{Runtime Evaluation}
Finally, we present the result of runtime evaluation for all cases mentioned above, as shown in TABLE \ref{table2}. The configuration of the PC platform is Intel Core i7-10700F CPU @2.90GHz. 
Our method computes the full trajectories for 10 quadrotors with formation change within 0.39s, which is efficient for such a complex task.  It can be seen that the trajectory optimization takes the most expensive computational cost, while growing cubically with the size of the state \cite{RN2}. There is a rapid increase in runtime from 6 quadrotors to 10 quadrotors, because their scenario configurations are totally different (a four-fold map and more complex transitions), not just a growth in the dimension of states. Compared with GPMP2, we just add two new kinds of unary factors, and the useful sparsity is still available. Therefore, the description of scalability in \cite{RN2} is still valid for our method. As for the global planning, formation planning and time allocation are independent of the number of robots, while task assignment is also insensitive to the scale of the quadrotor team.

\section{CONCLUSIONS}

In this paper, we propose a novel trajectory generation framework for the multi-robot formation. In the global planning part, we present a formation planning method and a task assignment approach for rectangular formations, to compute the expected formation sequence and the corresponding assignment scheme, which provides strong priors to encourage the following trajectory optimization to converge to the optimal solution.  Our trajectory optimization part is built upon GPMP2 by adding new constraints specific to multi-robot formation cases, which has high efficiency and good scalability by exploiting the sparsity brought by GPs. Both simulations and real-world experiments, where a team of quadrotors move through a width-varying area, show the efficiency and feasibility of our method. Additionally, a fast incremental replanning approach is implemented to illustrate the possibility of online operation. 



\begin{thebibliography}{99}
\bibitem{RN11}
J.~Alonso-Mora, S.~Baker, and D.~Rus, ``Multi-robot formation control and
  object transport in dynamic environments via constrained optimization,''
  \emph{The International Journal of Robotics Research}, vol.~36, no.~9, pp.
  1000--1021, 2017.

\bibitem{RN6}
W.~Hönig, J.~A. Preiss, T.~K.~S. Kumar, G.~S. Sukhatme, and N.~Ayanian,
  ``Trajectory planning for quadrotor swarms,'' \emph{IEEE Transactions on
  Robotics}, vol.~34, no.~4, pp. 856--869, 2018.

\bibitem{RN5}
C.~E. Luis, M.~Vukosavljev, and A.~P. Schoellig, ``Online trajectory generation
  with distributed model predictive control for multi-robot motion planning,''
  \emph{IEEE Robotics and Automation Letters}, vol.~5, no.~2, pp. 604--611,
  2020.

\bibitem{jackson2020scalable}
B.~E. Jackson, T.~A. Howell, K.~Shah, M.~Schwager, and Z.~Manchester,
  ``Scalable cooperative transport of cable-suspended loads with uavs using
  distributed trajectory optimization,'' \emph{IEEE Robotics and Automation
  Letters}, vol.~5, no.~2, pp. 3368--3374, 2020.

\bibitem{RN14}
D.~Mellinger, A.~Kushleyev, and V.~Kumar, ``Mixed-integer quadratic program
  trajectory generation for heterogeneous quadrotor teams,'' in \emph{2012 IEEE
  International Conference on Robotics and Automation (ICRA)}.\hskip 1em plus
  0.5em minus 0.4em\relax IEEE, 2012, pp. 477--483.

\bibitem{RN13}
J.~Alonso-Mora, S.~Baker, and D.~Rus, ``Multi-robot navigation in formation via
  sequential convex programming,'' in \emph{2015 IEEE/RSJ International
  Conference on Intelligent Robots and Systems (IROS)}.\hskip 1em plus 0.5em
  minus 0.4em\relax IEEE, 2015, pp. 4634--4641.

\bibitem{RN2}
M.~Mukadam, J.~Dong, X.~Yan, F.~Dellaert, and B.~Boots, ``{Continuous-time
  Gaussian process motion planning via probabilistic inference},'' \emph{The
  International Journal of Robotics Research}, vol.~37, no.~11, pp. 1319--1340,
  2018.

\bibitem{balch1998behavior}
T.~Balch and R.~C. Arkin, ``Behavior-based formation control for multirobot
  teams,'' \emph{IEEE Transactions on Robotics and Automation}, vol.~14, no.~6,
  pp. 926--939, 1998.

\bibitem{balch2000social}
T.~Balch and M.~Hybinette, ``Social potentials for scalable multi-robot
  formations,'' in \emph{Proceedings 2000 ICRA. Millennium Conference. IEEE
  International Conference on Robotics and Automation. Symposia Proceedings},
  vol.~1.\hskip 1em plus 0.5em minus 0.4em\relax IEEE, 2000, pp. 73--80.

\bibitem{zhou2018agile}
D.~Zhou, Z.~Wang, and M.~Schwager, ``Agile coordination and assistive collision
  avoidance for quadrotor swarms using virtual structures,'' \emph{IEEE
  Transactions on Robotics}, vol.~34, no.~4, pp. 916--923, 2018.

\bibitem{RN10}
W.~Ren and N.~Sorensen, ``Distributed coordination architecture for multi-robot
  formation control,'' \emph{Robotics and Autonomous Systems}, vol.~56, no.~4,
  pp. 324--333, 2008.

\bibitem{dunbar2002model}
W.~B. Dunbar and R.~M. Murray, ``Model predictive control of coordinated
  multi-vehicle formations,'' in \emph{Proceedings of the 41st IEEE Conference
  on Decision and Control}, vol.~4, 2002, pp. 4631--4636.

\bibitem{Liu2018}
S.~Liu, ``{Motion planning for micro aerial vehicles},'' Ph.D. dissertation,
  University of Pennsylvania, 2018.

\bibitem{RN25}
J.~Park, J.~Kim, I.~Jang, and H.~J. Kim, ``Efficient multi-agent trajectory
  planning with feasibility guarantee using relative bernstein polynomial,'' in
  \emph{2020 IEEE International Conference on Robotics and Automation
  (ICRA)}.\hskip 1em plus 0.5em minus 0.4em\relax IEEE, 2020, pp. 434--440.

\bibitem{RN7}
J.~Dong, M.~Mukadam, F.~Dellaert, and B.~Boots, ``{Motion Planning as
  Probabilistic Inference using Gaussian Processes and Factor Graphs},'' in
  \emph{Robotics: Science and Systems}, vol.~12, 2016.

\bibitem{RN17}
T.~D. Barfoot, C.~H. Tong, and S.~Särkkä, ``{Batch Continuous-Time Trajectory
  Estimation as Exactly Sparse Gaussian Process Regression},'' in
  \emph{Robotics: Science and Systems}, vol.~10, 2014.

\bibitem{RN20}
S.~Sarkka, A.~Solin, and J.~Hartikainen, ``{Spatiotemporal learning via
  infinite-dimensional Bayesian filtering and smoothing: A look at Gaussian
  process regression through Kalman filtering},'' \emph{IEEE Signal Processing
  Magazine}, vol.~30, no.~4, pp. 51--61, 2013.

\bibitem{RN21}
X.~Yan, V.~Indelman, and B.~Boots, ``{Incremental sparse GP regression for
  continuous-time trajectory estimation and mapping},'' \emph{Robotics and
  Autonomous Systems}, vol.~87, pp. 120--132, 2017.

\bibitem{RN19}
M.~Kaess, V.~Ila, R.~Roberts, and F.~Dellaert, ``{The Bayes tree: An
  algorithmic foundation for probabilistic robot mapping},'' in
  \emph{Algorithmic Foundations of Robotics IX}.\hskip 1em plus 0.5em minus
  0.4em\relax Springer, 2010, pp. 157--173.

\bibitem{RN18}
M.~Kaess, H.~Johannsson, R.~Roberts, V.~Ila, J.~J. Leonard, and F.~Dellaert,
  ``{iSAM2: Incremental smoothing and mapping using the Bayes tree},''
  \emph{The International Journal of Robotics Research}, vol.~31, no.~2, pp.
  216--235, 2012.

\bibitem{RN22}
F.~Dellaert, ``{Factor graphs and GTSAM: A hands-on introduction},'' Georgia
  Institute of Technology, Tech. Rep., 2012.

\bibitem{RN23}
M.~Zucker, N.~Ratliff, A.~D. Dragan, M.~Pivtoraiko, M.~Klingensmith, C.~M.
  Dellin, J.~A. Bagnell, and S.~S. Srinivasa, ``{CHOMP: Covariant hamiltonian
  optimization for motion planning},'' \emph{The International Journal of
  Robotics Research}, vol.~32, no. 9-10, pp. 1164--1193, 2013.

\bibitem{RN26}
J.~A. Preiss, W.~Honig, G.~S. Sukhatme, and N.~Ayanian, ``Crazyswarm: A large
  nano-quadcopter swarm,'' in \emph{2017 IEEE International Conference on
  Robotics and Automation (ICRA)}.\hskip 1em plus 0.5em minus 0.4em\relax IEEE,
  2017, pp. 3299--3304.
\end{thebibliography}
\end{document}